# Learning by Observation of Agent Software Images

**Paulo Costa**                                                    PAULO.COSTA@ISCTE.PT
**Luis Botelho**                                                    LUIS.BOTELHO@ISCTE.PT
*Instituto de Telecomunicações*
*ISCTE-Instituto Universitario de Lisboa*
*Avenida das Forças Armadas, 1649-026 Lisbon, Portugal.*

## Abstract

Learning by observation can be of key importance whenever agents sharing similar features want to learn from each other. This paper presents an agent architecture that enables software agents to learn by direct observation of the actions executed by expert agents while they are performing a task. This is possible because the proposed architecture displays information that is essential for observation, making it possible for software agents to observe each other.

The agent architecture supports a learning process that covers all aspects of learning by observation, such as discovering and observing experts, learning from the observed data, applying the acquired knowledge and evaluating the agent's progress. The evaluation provides control over the decision to obtain new knowledge or apply the acquired knowledge to new problems.

We combine two methods for learning from the observed information. The first one, the recall method, uses the sequence on which the actions were observed to solve new problems. The second one, the classification method, categorizes the information in the observed data and determines to which set of categories the new problems belong.

Results show that agents are able to learn in conditions where common supervised learning algorithms fail, such as when agents do not know the results of their actions *a priori* or when not all the effects of the actions are visible. The results also show that our approach provides better results than other learning methods since it requires shorter learning periods.

## 1. Introduction

This paper describes the important aspects of our approach for enabling software agents to learn control mechanisms by directly observing the actions of an expert agent while it is performing a task. It shows the innovations of our proposal for an agent software image (Costa & Botelho, 2011) and presents a complete learning by observation process, following previous work on this subject (Costa & Botelho, 2012). The paper also presents the results of our approach in two different scenarios (see section 5). The learning method used by our approach is usually known in the artificial intelligence community as learning by observation, imitation learning, learning from demonstration, programming by demonstration, learning by watching or learning by showing. For consistency, learning by observation will be used from here on.

Learning by observation is one of the least common and most complex forms of learning amongst animals because it is singular to humans and to a strict number of superior mammals. It is also one of the most powerful socialisation mechanisms (Ramachandran, 2003; Bandura, 1977; Meunier, Monfardini, & Boussaoud, 2007). Research in neurology and psychology shows that learning by observation may well be one of the causes of the exponential growth of human technologies in the last centuries. Unlike natural selection, observation allows these capabilities to spread amongst individuals within the same generation (Ramachandran, 2000).

Section 2 shows that learning by observation is already being used in robotic agents, which proves the applicability of this learning technique in artificial systems. However, progress in learning by observation is limited to robotics because software agents cannot observe one another in the same way tangible entities can be observed. Learning by observation is useful for software agents because it provides a more direct approach to the problem to solve when compared with other techniques where agents learn from experience, such as reinforcement learning. Instead of spending time testing several hypotheses, artificial agents acquire the knowledge from an expert by directly observing its actions while it performs a task. This allows the artificial agents to directly know which actions are necessary to perform a specific task (Argall, Chernova, Veloso, & Browning, 2009; Chernova, 2009).

Being able to observe an expert agent performing its actions (as opposed to merely rely on the observation of their effects) can be advantageous in situations where the effects of those actions are not directly visible in the environment (for example, agent communication, manipulation of software objects). Even when part of the effect of the actions is visible, directly observing the actions is still advantageous when the same effects could be achieved by different alternative actions but using one of them is clearly better than using others (for example, using a set of sums instead of a simple multiplication). Observing the actions performed by an expert is also advantageous when the representation of world states requires too much memory making it impossible to build large enough training sets (for example, web agents, manipulation of large databases) and especially when the agent does not know the effects of its actions *a priori* (for example, executing or invoking a software program or API).

The application of learning by observation in software agents is ideal for societies where agents share common features but have their own internal representation methods (for example, integration of legacy systems). Without common internal representation methods, directly transferring knowledge between agents is impossible or, at least very difficult. Learning by observation makes it possible to learn without the need of common internal representations because each agent makes its own interpretations of what it is observing.

The advantages described above provided the motivation for developing our approach to learning by observation in software agents. The major contribution of our approach is the definition of the whole learning process, which includes the discovery, observation, storage and interpretation of the observed data, the application of the acquired knowledge and a continuous internal evaluation (see section 4). Agents that learn by observation are capable of solving tasks in the same conditions as they were performed by the observed experts. They are also capable of performing a similar task when facing different conditions. The scenarios described in section 5 show these possibilities.

As section 2 shows, to the best of our knowledge, the only approach to learning by observation in software agents is the one presented by Machado and Botelho (2006). However, in Machado and Botelho's proposal the agents were only capable of learning vocabulary whereas in the work described in this paper the agents are capable of learning control mechanisms, which requires developing a different kind of learning algorithm. In addition, our approach also introduces several improvements to the software image (see section 3) initially proposed by Machado and Botelho (2006).

The software image provides software agents with an accessible representation of their constituents and capabilities, the static image, and of the actions they perform, the dynamic image. Our approach heavily relies on an agent architecture with software image. Agents with software image can be "seen" (consulted) by other agents and even by themselves. It is the software image that allows software agents to learn by observation because, in software environments, the information required for learning by observation is not automatically visible in the agent's body as it happens in the physical world (Quick, Dautenhahn, Nehaniv, & Roberts, 2000). To avoid misconceptions with the observation of tangible entities in the physical world, the act of observing a software agent will be defined as:

> Reading meta-data about an agent's constituents, its actions and the conditions holding when the actions were executed, without a direct intervention of the observed agent.

This allows the observed agent (the expert) to have only a passive role in the observation process. As with learning by observation in humans and superior mammals, it is the apprentice who takes action to obtain new knowledge, and does it without interfering with the expert agent. In spite of the similarities with Machado and Botelho's (2006) approach, our approach introduced several improvements to the software image, in particular:

- The inclusion of the agent sensors in the static image, in addition to the visible attributes, the actuators and the actions from Machado and Botelho's approach. This improvement provides a better description of the agent and allows other agents to know the way that agent understands the world, as explained in section 3.1.

- The dynamic image, besides displaying the executed action as in Machado and Botelho's approach, also displays information on the conditions holding before the action was executed, that is, the state of the environment, as perceived by the agent sensors, and the instances of the visible attributes. This improvement enables our agents to provide data for observation that is similar to the training sequences used for training machine learning algorithms (condition-action pairs), as explained in section 3.1.

- Our software image displays historical information about the actions executed in the past and about the conditions holding before each action was executed. This enables our agents to acquire a great amount of knowledge at the beginning of the observation, as section 3.2 shows.

- Our software image also uses ontologies to hold the knowledge on the designations of the different kinds of sensors, visible attributes, actions and tasks. This improvement allows agents that follow the same ontology to use the same designations for the same kinds of sensors, visible attributes, actions and tasks, as section 3.2 shows.

The contributions of our approach are not restricted to the agent software image. The other important contributions are the complete approach to learning by observation, in particular:

- The discovery of experts which provides the agent with the necessary tools to discover, by itself, expert agents from which it is possible to learn (see section 4.1).

- The definition of two learning algorithms, which are used to convert the observed information into mechanisms for choosing the actions to perform in each condition - the recall algorithm and the classification algorithm. The recall algorithm was totally developed for our approach and uses the sequence on which the expert actions were observed to choose the actions to perform. The classification algorithm is an adaptation of the existing KStar algorithm (Cleary & Trigg, 1995). It categorizes the information in the observed data and determines which actions to perform according to the categories of the new problems (see section 4.2).

- The definition of an internal evaluation which provides the agent with a measure of the confidence on its knowledge. Depending on this confidence, the agent can be in one of two states of the learning process: the learning state or the execution state (see section 4). When agents are in the learning state, their only objective is to observe experts and acquire new knowledge from them. When agents are in the execution state, their only objective is to perform their task using the acquired knowledge. Switching between these two states depends on two configurable thresholds, the UpperConfidenceThreshold and the LowerConfidenceThreshold, as explained in section 4.3.

- The ability to mimic the mirror neurons, which allows the agent to use the same mechanisms to propose actions both when learning and when using the acquired knowledge. This allows the agent to directly associate the observed actions with its own actions. It also allows the agent to propose actions for the conditions faced by the observed expert and determine if the agent is capable of proposing the same action as the observed expert.

The motivation that drives an agent to observe an expert is only partially covered by our approach. For simplification purposes, all apprentice agents have the observation of experts as their top priority. The apprentice agents are equipped with a specialized sensor that focuses their attention on observing an expert. The internal evaluation allows agents to decide whether or not they need more knowledge, providing them with the necessary motivation for observing experts, the only way they know how to obtain new knowledge.

Another simplification relates with determining if the experts to observe are performing relevant actions for the task to learn, which is also usually disregarded by the approaches for learning by observation. In normal circumstances, although an expert has the same features as the apprentice, it does not necessarily mean that it is performing the actions that are necessary for the apprentice to learn (by observation) how to perform a specific task. Like in all the surveyed approaches, the experts developed for the application scenarios in section 5 are prepared to only execute the actions that are necessary for the task to be learnt.

In section 5 the capabilities of our learning approach are tested in two different scenarios. The first scenario was especially conceived as a situation in which the majority of the effects of agent actions are not visible in the environment. Therefore, it will not be possible for a common machine learning solution to learn only from the effects of the actions. For a complete understanding of what is happening, it is necessary to observe the agent and its actions. The second scenario presents the mountain car problem as described by Sutton and Barto (1998). It is used to compare our learning approach with a reinforcement learning (RL) algorithm in a situation in which this kind of learning algorithm has already shown to be a good approach (Mitchell, 1997).

The results from the tests show that, with our approach, agents can correctly learn how to perform a task when the majority of the effects of agent actions are not visible in the environment (see section 5.1). In addition, when tested in situations where other learning methods may provide good results (see section 5.2), such as in a reinforcement learning scenario, results show that our approach is able to learn faster than the reinforcement learning approach. Besides learning faster, the agents using our learning method also require fewer actions to achieve the goal.

The following section presents a survey on research on the visual representation of agents and on learning by observation. Section 3 presents the improvements we have made on previous proposals regarding the software image. Section 4 describes the important aspects of the learning by observation process. Section 5 describes the test scenarios and experimental results. Finally section 6 presents conclusions and future work.

## 2. Literature Review

This section presents a survey of the approaches for learning by observation and for the visual representation of agents. It describes the important aspects of approaches related to learning by observation or that may contribute to solve the problems faced when applying learning by observation in software agents.

### 2.1 The Visible Representation of Software Agents

The literature overview regarding learning by observation shows that, with the exception of Machado and Botelho's (2006) approach, software agents are disregarded from the advances on learning by observation since they are usually related to robotics (Argall et al., 2009). Software agents are not able to distinguish themselves or other software agents from the remaining elements of a computer program because of the disembodied nature of software. Because of this, software agents are not able to observe each other as tangible entities would be observed (Etzioni, 1993; Quick et al., 2000).

Almost all software approaches for learning are constrained to observe the changes in the environment (the effects of agent actions) and the knowledge obtained to perform a task is limited to state change information (Quick et al., 2000; Argall et al., 2009). However, several authors have emphasized that not every action produces visible changes in the environment (Byrne, 1999; Dautenhahn & Nehaniv, 2002; Botelho & Figueiredo, 2004; Machado, 2006), thus, using state change information alone is not always a good option. In such cases, it is important to actually "see" what the expert agent is doing by observing its actions in addition to their effects. This allows the apprentice agents to know exactly what actions are necessary to perform a task, thus overcoming the problems that arise when the effects of those actions are not visible in the environment or when the agent does not know the effects of its actions *a priori* (Byrne, 1999; Dautenhahn & Nehaniv, 2002; Botelho & Figueiredo, 2004; Machado, 2006).

To be able to learn by observing the actions of other agents, the software agent needs some kind of accessible representation of its body that displays the necessary visible features (Mataric, 1997; Botelho & Figueiredo, 2004; Machado, 2006). Research in neurology reveals that the human brain also uses a representation of the human body in its activities (Ramachandran, 2003). This representation provides information on the body constituents and their possible actions and comes into existence in the initial stages of infant development. It is an important factor for learning by observation, since it allows children to acknowledge their bodies and capabilities. It also provides the ability to identify entities that are "similar to them" and thus may be worthwhile observing (Rao, Shon, & Meltzoff, 2004).

Despite these neurological findings, the literature review on embodiment and embodied cognition shows that, besides our approach (Costa & Botelho, 2011), only another one addresses the problem of creating such kind of accessible representation for software agents (Machado & Botelho, 2006). To the best of our knowledge, these are the only known approaches with a concrete proposal deploying a visible image for software agents, which is called the "visible software image" or simply the "software image". The literature review on learning by observation also provides no alternatives since all approaches, with the exception of the referred ones, apply exclusively to robotics (Argall et al., 2009).

Although Machado and Botelho's (2006) approach for the software image provides a description of the agent constituents and actions, it does not describe the kind of input the agent can collect from the software environment. Etzioni (1993) was one of the first authors to realize that the lack of a physical body was not an obstacle for the application of the principles of embodiment in software agents in the same way they are applied in robotics. For Etzioni, a software agent can be situated in the software environment in the same way as a robot is situated in the physical world, if it is characterized by what it can do (its actions) and also by the kind of inputs it is able to collect from the software environment (Etzioni, 1993).

For this reason, in our approach (Costa & Botelho, 2011), the software image provides software agents with a visible representation of their constituents, which includes their sensing and action capabilities, of the actions executed by the agent and the conditions holding when the agent decided to execute those actions. Our proposal also keeps an historical record of the actions performed by the agent and the conditions holding when the agent decided to execute those actions, for a limited amount of time.

## 2.2 Learning by Observation

The survey on learning by observation shows that one of the most important aspects of an approach to learning by observation is the learning algorithm. It defines how the knowledge, obtained from observation, is stored and how it is used, that is, how the agent proposes actions to execute when facing new problems (Argall et al., 2009). One possibility for the learning algorithm is to follow the same sequence of actions as the expert, which requires the agent to store the sequence on which it has observed the actions performed by the expert.

This possibility for the learning algorithm, named sequencing, is one of the most commonly used in learning by observation approaches (Argall et al., 2009). It is also closely related to sequence learning in humans since it handles the same kind of problems, such as predicting the elements of a sequence based on the preceding element, finding the natural order of the elements in a sequence and selecting a sequence of actions to achieve a goal (Clegg, DiGirolamo, & Keele, 1998; Sun, 2001). The best way to maintain the temporal relations between the elements in the sequence consists of using the properties of the data structure where the sequences are stored (Byrne, 1999; Heyes & Ray, 2000; Kulic, Ott, Lee, Ishikawa, & Nakamura, 2011; Billing, Hellström, & Janlert, 2011).

Linear structures such as lists and vectors are the most commonly used (Byrne, 1999; Heyes & Ray, 2000). However these linear structures lack the ability to represent alternatives, which is an important aspect of sequential learning that opens the possibility of making choices inside the sequence (Sun, 2001). The representation of alternative paths is essential for representing different approaches to perform the same task, that is, when the same objective can be reached by different sequences of actions. To hold this information, each different approach needs to be stored as an alternative sequence of actions. Tree structures are ideal for these situations. Each element of the sequence is represented as a node in the tree and the following element is chosen from one of the branches of that node (Kulic et al., 2011).

Sequencing is best suited for situations where agents face the same conditions (the same sequence of states of the environment and internal states) as the observed experts, which often means that the agent is following the expert or that it is possible to cover all the possible combinations of the conditions in the time the agent is observing.

Other possibilities for the learning algorithm are to generalize the acquired knowledge or to use analogies between the acquired knowledge and the new problems. This allows the agent to face future conditions that have never been observed, because in a real world situation it is almost always impossible to observe all possible conditions (Argall et al., 2009; Sullivan, 2011). Unlike in sequencing, there is no specific sequence of actions to follow. The agent determines what actions it should perform supported exclusively by the conditions.

One way of generalizing the acquired knowledge is using the observed conditions and actions to train neural networks, as described by Billard and Hayes (1999). Other hypothesis consist of using the observed conditions and actions to feed statistical approaches such as Bayesian algorithms and Hidden Markov models (HMMs) (Rao et al., 2004; Hajimirsadeghi & Ahmadabadi, 2010). The conditions and actions can also be used to train supervised learning algorithms, such as the classification algorithms (Argall et al., 2009; Chernova, 2009; Sullivan, 2011).

Given the advantages of the sequencing and of the generalization or analogy possibilities our approach to learning by observation presents a sequencing possibility, through the recall method of learning, and classification possibility, trough the classification method of learning. The two methods of learning are combined to increase the adaptability of the apprentice agents. The classification method allows the agent to extend its knowledge to conditions that have not been observed and recall method allows the agent to easily learn sequences of actions with different alternatives.

The survey on learning by observation also reveals that a learning by observation approach cannot be limited to the learning algorithm. In addition to the algorithm, it must also include the agent's motivation to learn, the discovery and observation of agents, the storage and interpretation of the information acquired in observation and the application of the newly acquired knowledge (Demiris & Hayes, 2002; Tan, 2012). One of the major flaws detected on the surveyed approaches, besides their focus on robot agents, was the fact that this global view is still missing (Tan, 2012). To the best of our knowledge, with the exception of our approach (Costa & Botelho, 2012), all approaches are focused on solving specific problems, and the solutions they provide are supported exclusively by the learning algorithm.

Demiris and Hayes (2002) provide a good starting point for building an approach that includes all aspects of learning by observation. Their approach presents a learning process that, excluding motivation, is consistent with Bandura's (1977) social learning theory, which approximates their approach to learning by observation in humans and superior mammals. The inclusion of an internal evaluation to Demiris and Hayes's (2002) approach provides the learning process with a simple motivation mechanism. The evaluation allows the agent to measure how its performance is affected both when it is consolidating the knowledge acquired from observation or when executing actions (Wood, 2008; Hajimirsadeghi & Ahmadabadi, 2010).

The agent can be intrinsically motivated to learn because it knows when it has acquired sufficient knowledge to perform a task on its own or because it detects that portions of its knowledge need improvement and thus require the agent to go back learning (Wood, 2008; Billing, Hellström, & Janlert, 2010). The ability to enhance the agent's knowledge through new observations is an important factor for learning by observation. According to Argall et al. (2009), one of the downsides of learning by observation is the fact that the agent's knowledge is limited to what it was able to observe. Using an evaluation stage, that operates when the agent is learning and when it is executing actions, provides the knowledge on when it is necessary to observe experts and when agents are ready to execute actions.

Several authors use specialized experts, or teachers, who monitor and reinforce the agent's actions (Sullivan, 2011; Hajimirsadeghi & Ahmadabadi, 2010; Chernova, 2009). They also measure the agent's performance and provide the necessary evaluation. Besides monitoring, the teachers can also take corrective measures like providing the appropriate actions for the faced conditions when the agent chooses incorrect actions (Hajimirsadeghi & Ahmadabadi, 2010). The teacher can also decide when the agent needs to acquire more knowledge (Sullivan, 2011).

The relation between the teacher and the agent can be extended by allowing them to communicate with each other. This allows the agent to request the teacher to perform a specific task (Chernova, 2009). However, this requires an additional effort in designing the expert agents since they need to communicate with the apprentice agents and teach them how to perform a task. This also requires apprentice agents to wait for the teacher to be available to communicate with them, which can extend the amount of time spent on learning. This does not happen when the expert plays a passive role in observation and, in addition, using teachers makes the approaches closer to learning by teaching, which goes beyond learning by observation.

A different approach for evaluation is mimicking the properties of mirror neurons. The mirror neurons are brain structures that exist in humans and superior mammals which are responsible for the emergence of learning by observation. They are involved in the tight coupling of perception and motor control, providing similar responses both when observing and when performing an activity. This allows the agents to "feel" like they are performing the actions they observe on an expert (Ramachandran, 2000), which greatly improves the easiness of identifying the observed actions, grounding them in the agent own actions.

Through this ability, the agent is able to propose actions, using its own mechanisms, for what it is observing, without effectively executing them. The proposed actions can be compared with those observed to determine if the agent is able to propose the same actions as the expert. The information provided by this comparison feeds the agent's internal confidence that it is able to propose the correct actions. In this case, the agent's internal confidence builds on the successes and failures of the actions that were previously proposed instead of the metrics on the current actions, provided by the learning algorithm, as in the work of Chernova (2009), and Billing et al. (2010).

Several approaches use specific structures, such as the forward models, to emulate the behaviour of mirror neurons (Rizzolatti, Fadiga, & Gallese, 1996; Demiris & Hayes, 2002; Maistros & Hayes, 2004; Rao et al., 2004; Lopes & Santos-Victor, 2007). However these structures are specifically designed for robots and use hardware inhibitors to prevent the robot actuators from executing the estimated actions. The main objective of these kind of structures is to create a distinction between the description of the action and its execution, that is, to create abstract representations of agent actions (Kulic et al., 2011). This solution is simpler and provides the agent with control over the execution of actions.

One of the characteristics of the reviewed approaches is that they usually focus on specific problems, which implies adaptations of some of their features whenever they are applied in new domains. Despite this problem, as shown in this section, the reviewed approaches provide important features that can be adapted to our approach to learning by observation.

## 3. The Software Image

This section presents a summary of the additional functionalities that our approach introduced to previous work regarding the software image (Machado & Botelho, 2006). The section describes these new functionalities, explains the reasons for their inclusion in the software image and how they are advantageous for learning by observation.

Our approach to learning by observation proposes a software image that allows software agents to learn by observing the actions of other agents. The act of observing a software agent does not imply the use of computer vision; instead the agents use specialized sensors that read meta-data about the observed agent. This meta-data is what we call the software image, which is defined by software objects and the relationships among them, as displayed in Figure 1. Despite the current version of the software image being primarily focused on learning by observation, we believe that this image can be useful for other purposes (for example, improve the agent's interaction with the surrounding environment through embodiment). Additional work will incrementally reveal the characteristics of a software image that is independent of any particular use.

Our proposal for the software image (Costa & Botelho, 2011) provides an accessible and domain independent description of the agent's constituents, the actions it executes and the conditions holding when it decided to execute those actions. This description is used by software agents that are interested in observing the represented agent for comparison with their own description, to check if both agents share the same capabilities, as described in section 4.1. Figure 1 shows a representation of the key elements of the software image. Like in Machado and Botelho's (2006) proposal, the elements of the agent software image are arranged in two categories, the static image and the dynamic image. The static image is immutable (does not change over time) and describes the constituents of the agent whereas the dynamic image changes with time and describes the activities of the agent.

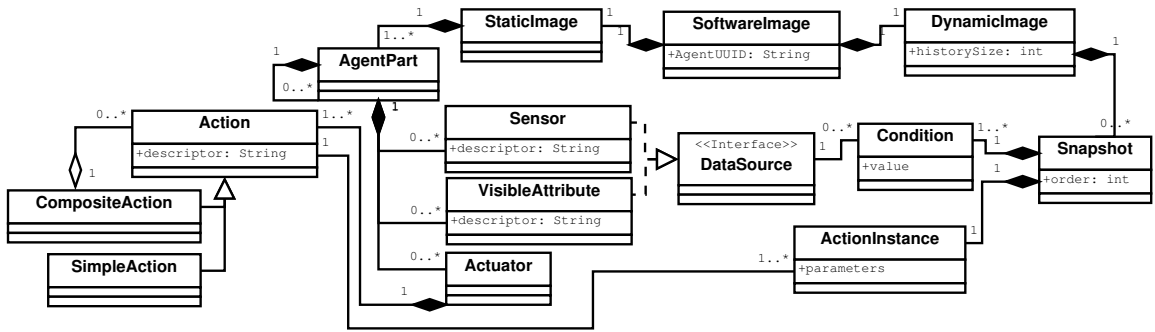

Figure 1: The class diagram of the software image

The improvements on Machado and Botelho's (2006) proposal for the software image consist of including the agent sensors in the description of its components (the static image), combining the information on the state of the software environment (provided by the agent sensors) with information on important aspects of the agent's internal state with the observed actions and enabling the representation of composite actions, that is, actions composed of sequences of simpler actions. Other important improvements include the ability to store historic data on the agent actions and on the conditions holding for those actions, and the use of an ontology to represent the knowledge on the agent sensors, actions and visible attributes, on the tasks to be accomplished and on the concepts and relationships that exist between those elements. Our proposal for the software image also defines a protocol for observing the snapshots.

The following sections describe these improvements on the software image in greater detail.

## 3.1 The Agent Sensors and the Snapshots of the Agent Activity

Machado and Botelho's (2006) version of the software image described software agents as a collection of parts with visible attributes and actuators. The actuators, on their turn, described a collection of actions that the agent was able to perform. However, it is also important to include the agent sensors in this description, especially when software agents only have access to part of the state of the environment, which is acquired by their sensors.

The information collected by the agent sensors represents the way the agent understands the world - the agent's perspective of the world. If the sensors are not included in the static image, the agents have no way of knowing if the experts they observe understand the world in the same way as them. The ability to understand the world in the same way as the expert is important for a better understanding of the reasons behind the expert's actions (Bandura, 1977; Ramachandran, 2000).

Given the importance of the agent sensors, our proposal for the software image includes them as constituents of the agent part (see Figure 1). This provides an accurate description of the agent and allows the agents to compare with others not only by the actions they can perform but also by the kind of information they can obtain from the environment.

On our proposal for the software image we also consider the conditions holding for an agent action as the state of the environment, as it is acquired by the agent sensors, and the instances of the agent's visible attributes (the important aspects of the agent's internal state) at the moment the agent selects that action. The information contained in the conditions depends exclusively on the information provided by the agent sensors and by the visible attributes. The agent sensors and visible attributes, as well as the type of information they provide, are defined when designing the agent. Section 5 shows an example of how the sensors and visible attributes are set for an agent and how this affects the conditions.

The conditions holding for an action play an important role in the information provided for observation in the dynamic image. Unlike Machado and Botelho's (2006) proposal where agents could only observe the action being currently performed, in our proposal, agents acquire snapshots of the activity of the observed agent. Each snapshot contains information on the executed action and on the conditions holding at the moment the agent decided to select it. This way, the information provided for observation is similar to the training sequences used for training machine learning algorithms (a sequence of condition-action pairs), which is an important aspect of the learning methods described in section 4.2.

In addition to including the conditions in the information provided for observation, the action provided in the snapshot can either be simple or composite (see Figure 1), that is, an action composed of a sequence of actions. This allows agents to handle sequences of actions in the same was as single actions, when observing.

## 3.2 Additional Innovations on the Software Image

This section describes additional innovations on our proposal for the software image. The most important innovation is the ability to store historic data. The historic data provides a limited amount of past snapshots. It allows observers to gather knowledge much faster, when compared with observing only the current action, because it is not necessary to wait for the agent to perform those actions. However, this innovation requires the conditions to hold the perspective of the agent executing the actions, that is, both the state of the environment and the instances of the visible attributes are provided as the agent perceives them. Without the agent's perspective it would be hard, if not even impossible, for the observer to know the conditions holding in the past.

Using the agent's perspective can be seen as a limitation when compared with using the observer's perspective when acquiring the conditions because it requires the observer agents to have the same kind of sensors and visible attributes as the expert agents they observe, so they can understand the information contained in the conditions (see section 4.1). However, it is not always ensured that the observer has a direct access to the environment of the observed agent, like for example, when the software environment of the observed agent is running on a distinct process. In these cases, it would be necessary to use complex communication mechanisms for the observer to get access to the information on the different process. This would not be necessary if the observer only used the information provided by the software image because of an additional mechanism, the SOFTWARE IMAGE INDEX described in section 4.1, which provides a shared repository where all registered software images can be easily accessed.

Another innovation of the software image is the use of an ontology to describe the knowledge on the agent sensors, visible attributes and actions, on the tasks to be accomplished and on the relationships between these elements. Using ontologies allows different agents that follow the same ontology to use the same designations for the same kinds of sensors, actions and visible attributes and for the same tasks. The ontology also enables specific kinds of sensors, actions and visible attributes to be associated to specific tasks, which allows the agents to know which elements are required to perform a task (see section 4.1).

Another important aspect of the ontology is the possibility of associating two different elements, which opens the possibility of translations between different ontologies. A meta-ontology, which we call the software image meta-ontology, was created to facilitate this translation. The meta-ontology defines the basic elements of the ontology and the possible relationships between those elements. Additional information on this subject will be presented in future work.

In addition to these innovations, our proposal also defines a new protocol for observing the snapshots in the software image. A functionality developed for the software image, the DYNAMIC IMAGE NOTIFICATION, allows the subscribed observers to receive notifications each time a new snapshot is created on the dynamic image of the observed agent. This allows the observers to know exactly when they have to observe, that is, when they can collect a new snapshot from the dynamic image of the observed agent. The following section explains the way agents learn by observing (acquiring information from the agent's software image) an expert.

## 4. Learning by Observation

This section summarizes the approach regarding the complete learning by observation process, following previous work on this subject (Costa & Botelho, 2012). It shows a new insight on the learning process and focuses on important aspects such as the process of discovering and observing experts and the methods of learning from the information provided by observation. The section also describes the agent's internal evaluation and how it affects the agent's behaviour.

The approach to learning by observation requires both the expert and the apprentice agent to have software image since it provides the means for comparing the agents and also the data for observation (see section 3). It presents a global view of the learning process which comprises six activities, presented in Figure 2, that not always happen in strict sequence. The agent may also be in one of two states regarding the learning by observation process: the learning state and the execution state. In each of these states, the agent will have access to only a subset of the learning process activities. Figure 2 shows which activities are available to each state.

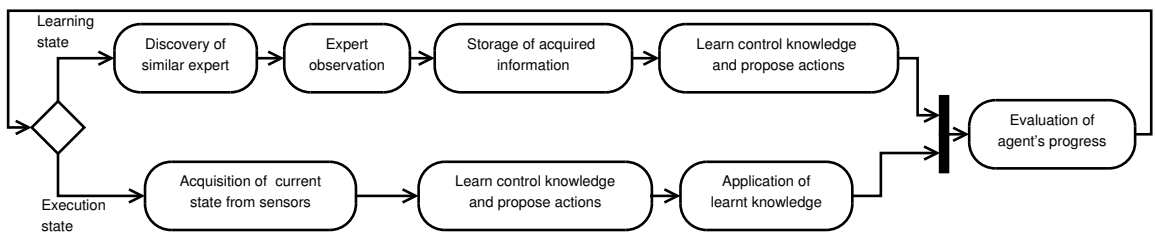

Figure 2: The activities of the learning process on each state

As Figure 2 shows, the possible activities of the learning state concern discovering and observing the expert, retaining the information acquired in observation, learning control knowledge to propose actions and evaluating the proposed actions. The possible activities of the execution state concern acquiring the current state of the environment (from the agent sensors), learning control knowledge to propose actions, executing the proposed actions and evaluating the executed actions. Figure 2 also shows that, as result of mimicking the properties of the mirror neurons, agents are capable of proposing actions both when learning and when executing actions (see section 4.2).

The approach provides two distinct methods of learning from the information acquired by observation, the recall and the classification methods, that where specifically developed for the approach. They were inspired on the two most used algorithms for learning by observation (see section 2.2). The two methods are combined to present a single solution for apprentice agents (see section 4.2), which increases the agent's ability to adapt to different situations. The agents are able to perform the observed task when facing the same conditions as the experts and also a similar task when facing different conditions.

The internal evaluation, described in section 4.3, is one of the most important activities in the learning process because it monitors the agent's ability to propose the correct actions over time. The result of this monitoring is a measure of the agent confidence on the learnt knowledge, the confidence value. The agent will be on the learning or on the execution state depending on its confidence regarding the acquired knowledge.

The following sections describe important aspects of the approach to learning by observation and of the learning process such as how experts are discovered, how the agent learns from the information acquired in observation and how the internal evaluation works.

## 4.1 Discovering and Observing Expert Agents

The software image, described in section 3, addresses the problem of providing information that is necessary for observation in a way that is universally accessible to software agents. With the software image, apprentice software agents can compare themselves with the discovered experts and collect information about the actions executed by the expert and about the conditions holding at the time the expert decided to execute those actions. A discovery service, the SOFTWARE IMAGE INDEX, was developed to facilitate the discovery of expert agents. It allows software agents to register their software images in a shared repository so that other agents are able to find them. The agents use this service to discover the software images of the expert agents.

Before starting to observe an expert, the agent must know if it is potentially possible to learn by observing that particular expert. When it is not possible to determine what agent structures are necessary for the task to learn (see section 3.2), the agents follow Bandura's (1977) social learning theory and learn by observing a similar expert, that is, an expert whose static image has the same structure and the same instances of the atomic elements as the agent's static image. The agent uses the comparison functionalities of the software image described by Costa and Botelho (2011) to compare its static image $SI_{agent}$ with the static image of the expert $SI_{expert}$. If both images match, $SI_{agent} \equiv SI_{expert}$, the expert can be observed because the agent is able to immediately recognize the actions and conditions on the snapshots (see section 3.1), solving most of the correspondence problems (Alissandrakis, Nehaniv, & Dautenhahn, 2002; Argall et al., 2009).

When the agent knows the task to learn $T$ and the structures and abilities that are necessary for that task, $si_T$ (see section 3.2), the concept of an expert from which it is potentially possible to learn is extended, allowing agents to observe an expert as long as the intersection of their software images contains those structures and abilities, $(SI_{agent} \cap SI_{expert}) \ni si_T$. This is enough to ensure that the apprentice agent is able to recognize all the conditions and activities on the expert snapshots that are necessary for learning a specific task. The agents determine which structures and abilities are necessary to perform a task through an ontology, as explained in section 3.2.

After discovering the expert to observe, the agent subscribes the expert's DYNAMIC IMAGE NOTIFICATION and acquires all the snapshots in its history record (see section 3.2). The notifications facilitate the process of observing the expert since they determine the ideal moment for the agent to observe, which is when a new snapshot is created in the expert's dynamic image. While the agent is acquiring the snapshots in the history record, the new snapshots created in the expert's dynamic image are also acquired and stored in a temporary memory.

The temporary memory functions as a buffer for observation because it allows the agent to handle snapshots at a different rate from which they are acquired. It also allows the agent to keep record of the expert's actions that might take place while it is reading the history. The snapshots stored in the temporary memory are only handled after the history snapshots are handled, which provides the agent with an uninterrupted sequence of snapshots from a moment in the past until the current moment.

When compared with other solutions such as searching the history record on demand, that is, to find relevant information for a particular problem, collecting all the information on the history record is a more efficient solution because it allows the agent to obtain a large set of experiences in a small amount of time. Acquiring all the history record not only ensures that the solution for a particular problem is found (if it really exists in the history) but also provides the agent with an increased amount of information that might be important to solve other problems in the future.

An important aspect of the approach to learning by observation is the preference for observing different experts. The process of discovering an expert and collecting snapshots from its dynamic image is referred to as the observation period. This process is cyclical meaning that the agent may go through several observation periods while learning. At the beginning of each observation period the agent is free to choose a different expert to observe, increasing the diversity of its knowledge because different experts might provide different points of view on the task to learn.

The different perspectives provided by the experts may increase the agent's knowledge with conditions that were never observed on previous agents or with a different approach for performing a task. The following section shows the way the agent increases its knowledge and handles these different approaches.

## 4.2 Learning from the Observed Snapshots

This section describes how the agent learns from the observed snapshots and how it uses the acquired knowledge to propose actions for a set of conditions. It describes how the agent holds the information contained in the snapshot in its memory. It presents the two methods for learning from the observed snapshots, the recall and classification methods, and explains how they are combined to present a single solution for proposing actions. The section also explains how the behaviour of mirror neurons is mimicked by the approach and why this is useful for learning.

The snapshots acquired from observation describe a relation between an optimal set of conditions $(C_1 \wedge \cdots \wedge Cn)$ and an action $A$, $C_1 \wedge C_2 \wedge \cdots \wedge C_n \to A$ (see section 3.1). This relation, which we call EXPERIENCE, is stored in the agent's memory which is held by a tree structure. Using a tree structure enables the sequence on which the snapshots were observed to be intrinsically preserved by the structure, as shown in the literature review in section 2. The tree structure also facilitates the consolidation of the agent's knowledge because it stores different approaches for the same task as alternative paths.

```
SET Newexp as the new experience to store in memory
SET Previous as the experience from the snapshot observed before
SET Stored to false
FOR each Exp <- experience in the agent's memory
    IF Exp has same conditions as Newexp
      AND Exp has same action as Newexp THEN
        PUT Exp in sub-tree of Previous
        SET Stored to true
        BREAK
    ENDIF
ENDFOR
IF Stored is false THEN
    ADD Newexp to the agent's memory
    PUT Newexp in sub-tree of Previous
ENDIF
```

Figure 3: The process storing experiences in the memory

Given that agents can observe different experts, the sequence on which the snapshots are observed is broken each time the agent starts observing another expert. When this happens, the snapshot observed before represents the activity of another expert and therefore cannot be followed by the new snapshots. To prevent the fragmentation and duplication (multiple instances of the same EXPERIENCE) of the agent's memory, the process of storing new knowledge in the agent's memory compares each new EXPERIENCE with the ones existing in memory before storing it. Figure 3 describes how the process takes place.

The process described in Figure 3 allows the agent to create new knowledge using the information that already exists in memory by merely creating new connections between the existing EXPERIENCES. A new EXPERIENCE is stored only when it does not exist in the agent's memory. The tree structure holding the agent's knowledge allows the agent's knowledge to be expressed as a decision tree, as shown in Figure 4, where each node of the tree is an EXPERIENCE. Depending on the number of branches starting from the node, each EXPERIENCE can be followed by one or more EXPERIENCE, which adds the possibility of choosing which sequence to follow and therefore provide different alternatives for executing a task.

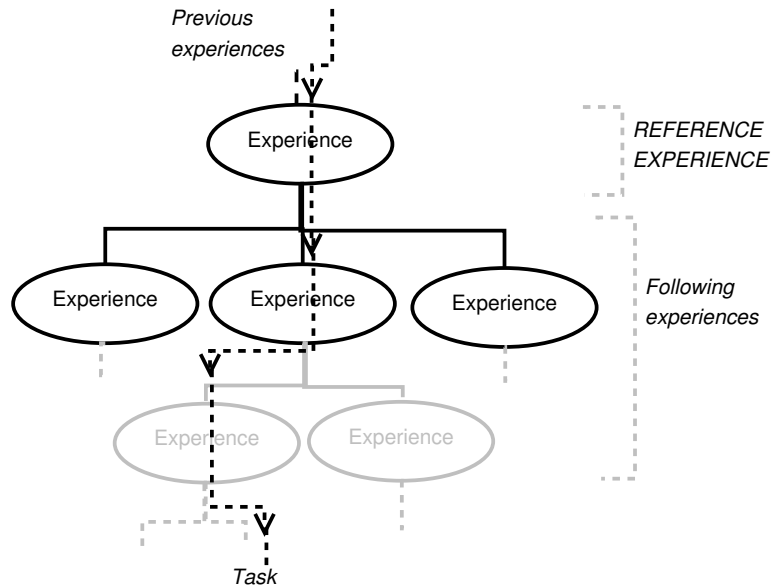

Figure 4: A representation of the tree structure in the agent's memory

The recall and the classification methods of learning use the information contained in the agent's memory (the tree of EXPERIENCES shown in Figure 4) to propose actions for given conditions, which we call the CURRENTCONDITIONS. The two methods distinguish themselves by the way they use the information in the agent's memory to propose the actions. The recall method proposes actions by determining the way the EXPERIENCES are connected with each other in the tree, whereas the classification method proposes actions by categorizing the individual EXPERIENCES and comparing those categories with the category of the current problem.

Both the recall and classification methods provide several possibilities of actions. To determine which of these actions is the best suited for execution, a RELIABILITY is associated to each proposed action. The RELIABILITY is calculated in each method and ranges from zero to one, inclusive. It determines how reliable an action is (zero not reliable; one fully reliable) for the method that proposed it.

The recall method proposes actions by following the connections between the EXPERIENCES in the agent's memory (see Figure 4). To propose actions, the method requires a reference to an EXPERIENCE in the agent's memory which indicates where to start following the connections. This reference, which we call the REFERENCEEXPERIENCE, represents the last action executed by the agent ACT and the conditions holding for that action COND. Figures 5 and 6 show the way the REFERENCEEXPERIENCE is discovered in the agent's memory.

```
FUNCTION discoverReferenceExperience(Cond,Act)
    SET Similarity to zero
    SET RefExp
    FOR each Exp <- experience in memory
        IF action from Exp is the same as Act THEN
            IF Exp has same conditions as Cond
                RETURN Exp
            ELSE
                SET Val as similarityBetween(conditions from Exp,Cond)
                IF Val is bigger than Similarity THEN
                    SET RefExp to Exp
                    SET Similarity to Val
                ENDIF
            ENDIF
        ENDIF
    ENDFOR
    RETURN Exp
ENDFUNCTION
```

Figure 5: Obtaining the REFERENCEEXPERIENCE from the agent's memory

As the process in Figure 5 shows, the REFERENCEEXPERIENCE is the experience in memory whose action is the same as the last action executed by the agent and whose conditions are the same as the conditions holding for the last executed action. Given that the agent might not be facing the same conditions as the expert, it is possible that the conditions holding for the last executed action are not found in the agent's memory. In these cases, the REFERENCEEXPERIENCE is the one with the same action who shares the largest number of similar conditions as calculated by the process shown in Figure 6.

After discovering the REFERENCEEXPERIENCE in the agent's memory, its tree subset (the set of tree branches) is retrieved. The tree subset represents the choices of the EXPERIENCES that follow the REFERENCEEXPERIENCE (see Figure 4). The actions from those EXPERIENCES represent the several action possibilities to be proposed by the recall method. The RELIABILITY of each action is obtained from the similarity between the conditions holding for the action (in the experience) and the given CURRENTCONDITIONS as calculated by the process shown in Figure 6. The highest RELIABILITY is given to the action whose conditions most resemble the CURRENTCONDITIONS.

Unlike the recall method, the classification method does not require a reference to an experience in the agent's memory to propose actions. The proposed actions are provided by an adaptation of a classification algorithm that is trained with the EXPERIENCES in the

```
REQUIRE CondA to have the same size as CondB
FUNCTION similarityBetween(CondA,CondB)
    SET Sum to zero
    SET Size as length of set of CondA
    FOR each C1 <- condition in Exp
        Inner <- (FOR each C2 <- condition in Cond)
            IF C1 equals C2 THEN
                ADD one to Sum
                BREAK Inner
            ENDIF
        ENDFOR
    ENDFOR
    RETURN (Sum / Size)
ENDFUNCTION
```

Figure 6: Obtaining the similarity between two sets of conditions

agent's memory. Previous experiments (Costa & Botelho, 2012) revealed that the most suited algorithms for the classification method are the KStar from Cleary and Trigg (1995) and the NNGE (Nearest Neighbour like algorithm using non-nested Generalized Exemplars from Martin, 1995). Since the latter consumes more resources the choice falls on the KStar algorithm.

The KStar algorithm was adapted to be able to propose representations of agent actions. The implementation of the KStar algorithm was modified to allow the EXPERIENCES in the agent's memory to be regarded as positive examples from which the proposed actions are deduced. The categorization capabilities were also enhanced to allow the conditions to define the differences between the classes. To propose actions, the classification method calculates the distances between the conditions to find the experiences whose conditions are closer to the CURRENTCONDITIONS. The KStar algorithm uses entropy to measure the distances between two conditions (Cleary & Trigg, 1995).

The RELIABILITY of the actions proposed by the classification method is directly associated with how much the conditions holding for that action are close to the CURRENTCONDITIONS. Once again, entropy is used to measure this distance. For example, if the conditions holding for a proposed action are identical to the CURRENTCONDITIONS the RELIABILITY of that action is 1, that is, the action is fully reliable from the standpoint of the classification method.

The recall and classification methods are combined in a single solution. Agents use both methods to propose actions and choose the best one for the current situation. To help deciding which action is best for the current situation, both the recall and the classification methods are associated to a WEIGHTFACTOR, whose initial and minimum value is zero. The WEIGHTFACTOR determines which of the methods is capable of proposing the most suitable action for the current situation. The RELIABILITY of the proposed action is combined with the WEIGHTFACTOR of the method that proposed it, which results in the FINALRELIABILITY ($finalReliability = reliability \times weightFactor$). The action with the highest FINALRELIABILITY is the best one for the current situation.

The WEIGHTFACTORS change each time the agent's capacity of proposing actions is evaluated (see section 4.3). The WEIGHTFACTOR of a method increases if the action with the highest RELIABILITY proposed by that method is proven to be an appropriate choice by the internal evaluation (see section 4.3). If the evaluation determines that the action was not appropriate, the WEIGHTFACTOR of the method decreases. As explained in section 4.3, the amount on which the WEIGHTFACTOR increases or decreases depends on the value of the RELIABILITY of the proposed action.

The recall and classification methods of learning propose actions when the agent is in the learning state and in the execution state. This is possible because the actions proposed by the methods are simply a representation of the agent actions, that is, the proposed actions are not automatically executed. The agent has the control over which action is going to be actually executed. This control allows the agent to behave in the same exact way both when observing (in the learning state) and when preparing to execute its own actions (the execution state), which in a sense is similar to what happens with the mirror neurons.

The ability to propose actions in the learning state allows the agent to experience the actions it observes as if it was preparing to perform them, by proposing actions for the conditions in the observed snapshots. This is advantageous since it allows the agent to realize if it is capable (or not) of making the same decisions as the expert. It is also useful for the agent's internal evaluation to decide if the agent has acquired sufficient knowledge to change to the execution state, as explained in the following section.

## 4.3 The Agent's Internal Evaluation

This section describes the agent's internal evaluation. It shows the way evaluation operates and how it affects the transition between the two states of the learning process, the learning and the execution state. It describes the way the agent's confidence is updated and how it is related to the agent's ability to propose appropriate actions.

The evaluation is a transversal process that covers both the learning and the execution state. The main purpose of evaluation is to ensure that the agent's knowledge is appropriate for mastering a task, which influences the agent's internal confidence. The agent's internal confidence expresses the successes and failures in proposing the appropriate actions for the faced conditions. Depending on the value of the internal confidence the agent may be in the learning state or in the execution state of the learning process. Two configurable thresholds, the UpperConfidenceThreshold and the LowerConfidenceThreshold determine the values at which the agent changes to the learning or to the execution state. Section 5.1 shows how the values selected for these thresholds influence the agent's capabilities. Figure 7 shows how the thresholds affect the transition between the two states of the learning process.

As Figure 7 shows, when the agent's confidence goes over the UpperConfidenceThreshold the agent is confident enough on its knowledge and switches to the execution state where it executes the actions it proposes (see section 4.2) for the conditions provided by its sensors and by the relevant aspects of its internal state (the visible attributes). When the confidence goes under the LowerConfidenceThreshold the agent stops being confident on its knowledge and switches to the learning state where it acquires more knowledge by observing experts. The value of the internal confidence is affected by the agent's capacity to propose the appropriate actions whether it is observing (in the learning state) or preparing to execute those actions (in the execution state).

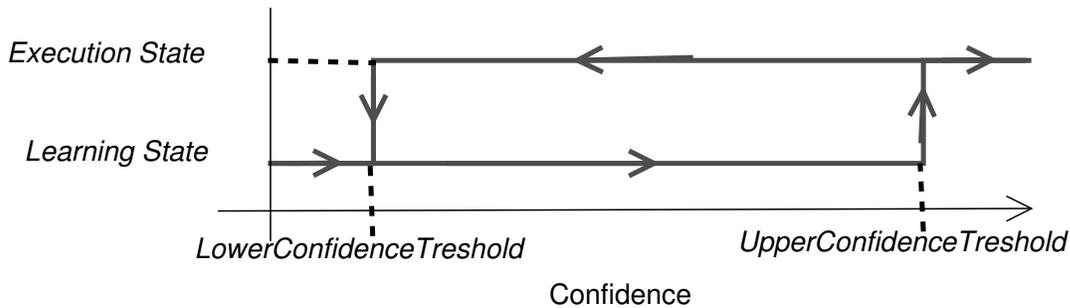

Figure 7: The influence of the confidence thresholds in state transition

The agent's internal evaluation is constantly testing the agent's capacity to propose the correct actions, both when the agent is in the learning state and in the execution state. In the learning state, the agent is tested for its ability to propose actions for the conditions in the observed snapshots. The agent's confidence increases when the best action it proposes (the one with the highest FINALRELIABILITY as explained in section 4.2) is the same as the action observed in the snapshot, otherwise the confidence decreases. The amount on which the confidence increases or decreases depends on the RELIABILITY of the best action proposed. For example, if $A$ is the action in the snapshot, $B$ is the best action proposed (with a RELIABILITY of 0.8) and $A \neq B$, the agent's confidence decreases 0.8.

Using the RELIABILITY of the best proposed action as a factor for increasing or decreasing the agent's confidence is the simplest way of including the confidence that the methods of learning have on the actions they propose (see section 4.2) in the calculation of the agent's internal confidence. The way the internal confidence is updated gives more importance to the actions proposed with a high RELIABILITY, that is, when the learning methods have a high confidence on the actions they propose. In these cases, if the actions are not correct the penalization in the internal confidence should be larger than when the methods have low confidence on the actions they propose because it means the agent learnt something wrong.

In the execution state, the agent is tested for its ability to execute the correct action for the faced conditions, which reflect the agent's perception from its sensors and the important aspects of its internal state (see section 3.1). As section 4.2 shows, when in the execution state the agent selects the action with the highest FINALRELIABILITY, from those proposed by the recall and classification methods, to be executed. A simple monitoring detects if there was any problem that prevented the correct execution of that action. Whenever a problem is detected the agent's confidence decreases. Once again the amount on which the confidence decreases depends on the RELIABILITY of the executed action (see section 4.2).

The simple monitoring of the execution of the actions has some limitations because even though there are no problems when executing the actions it is not possible to ensure that the action was the most appropriate for the faced conditions. To overcome this problem, our approach allows the evaluation to receive external feedback on the executed actions from specialized experts, the teachers, or from other evaluators that are specific to the applica-tion domain. This possibility for evaluation approximates our approach to the paradigm of learning by teaching which goes beyond the scope of this paper and therefore will be presented in future work.

Another important feature of the agent's internal evaluation is the ability to force the agent to switch to the learning state independently of its confidence. This forced switching happens when the agent is faced with a certain amount of unfamiliar conditions, that is, conditions not found on the observed snapshots. The amount of unfamiliar conditions is controlled by a configurable threshold whose value depends on the application domain. As explained in section 3.1, the conditions consist of the information provided by the agent sensors and visible attributes. The conditions are familiar when the agent has observed an expert facing the same conditions and therefore knows exactly the correct action to propose.

If the conditions are not familiar, the agent has never observed an expert facing them which probably means that it does not know what actions to propose. Under these circumstances, the amount of unfamiliar conditions rises only when the evaluation determines that the executed action was inappropriate. To turn this forced switching in a measure of last resort, the amount of unfamiliar conditions resets each time the agent faces a familiar condition. Therefore, this amount represents the number of consecutive times the agent faces unfamiliar conditions.

After being forced to switch to the learning state, the agent's confidence decreases to a configurable reference value, the UNFAMILARCONFIDENCEREFERENCE, which has to be lower than the LOWERCONFIDENCETHRESHOLD to allow the agent to remain in the learning state for a while. The lower the value of UNFAMILARCONFIDENCEREFERENCE in relation to the LOWERCONFIDENCETHRESHOLD the longer the agent takes to change back to the execution state. This is the simplest solution to keep the agent in the learning state for some time.

Other solutions which take the unfamiliar conditions into account are more complex because they require the agent to find an expert that is facing those conditions. Since it is impossible to determine when the agent finds such experts, it would be necessary to develop complex mechanisms to allow the agent to return to the execution state after some time, thus preventing the agent from spending too much time learning. A similar behaviour can be achieved by reducing the confidence and letting the internal evaluation decide when to return to the execution state by testing the agent's knowledge while it is observing.

The role played by evaluation, namely making the agent return to the learning state, contributes to overcome one of the major problems faced by learning by observation, the fact that the agents are limited to performing only the actions they have observed. Forcing the agent to return to the learning state (and thus to observe other experts) gives the agent an opportunity to increase its knowledge since other experts may have different experiences that might provide the agent with new knowledge.

## 5. Experimental Results

This section describes two scenarios in which our approach was implemented. Since both scenarios are software implementations, errors in the acquisition of the snapshots or of the conditions make no sense and therefore are not considered in the experiments. The first scenario was designed to be appropriate for learning by observation because the majority of expert activity does not affect the state of the environment. The second scenario compares our approach with a reinforcement learning approach in a typical reinforcement learning experiment, the mountain car experiment from Sutton and Barto (1998). Through this scenario we present a direct comparison between the solutions provided by a reinforcement learning algorithm and by our approach to learning by observation. The statistical relevance of the data collected for the comparisons is guaranteed by the student's t-test.

The scenarios are software implementations where all developed agents have a software image. When apprentice software agents observe the experts they are effectively acquiring snapshots from the expert's software image. As a simplification, the software images of the apprentice agents have the same constituents as the software images of the experts they observe. In both scenarios the apprentice agents do not receive external feedback from teachers or other evaluators (see section 4.3) so that the results reflect a pure learning by observation approach.

Despite the process of discovering and identifying a similar expert being an important step for the approach to learning by observation, the results presented in this section do not disclose this process for the purpose of focusing the results on the benefits of learning by observation. In both scenarios, the apprentice agents use the software image for discovering, identifying and collecting the information that is necessary for observation from the experts. Previous work on the software image (Costa & Botelho, 2011) presents some results on these aspects.

The first scenario simulates an agent with a virtual hand that displays numbers in sign language. The numbers are provided to the agent by a software number generator which provides numbers between one and five. Each agent is associated with a single number generator and has no knowledge on the other number generators associated to the agents that participate in the simulation. Each time a new number is provided to the agent, the virtual hand is changed to display that number by means of sign language. The virtual hand is used only for communicating with users through a graphical display and it is not accessible to other agents.

However, the agent's virtual hand has an accessible representation of its current state in the agent's software image as a visible attribute (see section 3). The only way software agents can obtain information on the state of the agent's virtual hand is through the software image. The visible attribute represents the state of the visible hand as an object with five attributes each representing a finger on the hand. Each attribute can be in one of two states, UP or DOWN.

The expert agent designed for this scenario has a specialized part which holds the knowledge on the most efficient way of changing the virtual hand so that it represents the number provided by the number generator. For example, when the agent perceives the number one, if the virtual hand is showing the number two (the index and middle fingers are UP) it is only necessary to move the middle finger DOWN, whereas if the hand was showing the number four, it would be necessary to move the middle, ring and pinky fingers DOWN. The part perceives the numbers from the number generator through a specialized sensor. It has an actuator with five actions, one for each finger, that change the state of that finger.

To increase the complexity of the scenario, the number generators need to be reset from time to time or else they will stop generating new numbers. The number sources can either be in the Active state or in the Inactive state. The source stops providing numbers when it is Inactive. The reset changes the source back to the Active state. Because of this, the expert agent has another specialized part which holds the knowledge on when and how to refresh the random number generator. The state of the number generator is perceived by this specialized part through a sensor which indicates the state of the generator. The agent part has one actuator with a single action that resets the source.

The apprentice agent developed for this scenario has to learn how to master these two tasks, manipulating the agent's virtual hand to display the perceived numbers and resetting the source, by observing experts with the same constituents and capabilities. Like the experts, apprentice agents have two parts. One of them specializes in manipulating the virtual hand and has one sensor that perceives the numbers provided by the number generator, one visible attribute that displays the state of the virtual hand and one actuator with five actions that change the state of each finger.

The other part of the apprentice agent specializes in managing the source that provides the numbers and has one sensor that perceives the state of the number source and one actuator with a single action that resets the source. Given the description of the agent sensors and visible attributes, the conditions for this scenario (see section 3.1) consist of the number provided by the agent source, the state of the virtual hand and the state of the number source.

The second scenario is a software implementation of an agent that learns how to climb a mountain simulated by a sinusoidal wave. The agent must abide by the laws of physics to climb the mountain, and because it has no sufficient force, it will not be able to climb the mountain by going forward only, it needs to accelerate backwards and forwards to gain momentum. The goal of this scenario is to reach the top of the mountain, that is, the peak of the sinusoidal wave, taking the least number of decisions and travelling the smallest distance (up and down the mountain) as possible.

The experts, provided for this scenario, know the optimal way (the exact moment and direction they need to accelerate) to climb the mountain and reach its top. The experts perceive their current speed and direction and their location in the mountain through their sensors and use this information to decide on the direction which they should accelerate next. They can choose between accelerating forward, accelerating backward or not accelerating (which maintains their current speed).

The apprentice agents have to learn to decide which direction to accelerate according to their location, speed and direction. The scenario includes two kinds of apprentice agents because we are comparing two different learning methods. The first kind of apprentice uses a reinforcement learning algorithm (Q-Learning, implemented in the PIQLE tool from Comité, 2005) to provide the agent with the knowledge that is required to climb the mountain. The reinforcement learning agent is able to perform three actions, accelerate forward, accelerate backwards or maintain speed.

The reinforcement learning agent is configured with a learning rate $\alpha = 0.2$ and a discount factor/rate $\gamma = 0.9$, which are the settings that present the best results. The learning rate also decreases with time following a geometrical decay, which allows the agent to eventually stop learning after a period of time. A simple reward scheme is used for the reinforcement learning algorithm. The agent is only rewarded when it reaches the goal of this scenario, the top of the mountain. Any other reward schemes would require the use of some kind of supervisor to determine the ideal situations to apply the reinforcements. This would have changed the comparison we intended for this scenario, which is to compare learning by observation with a situation where apprentice agents need no supervisors. This is the case for the reinforcement learning agent with this simple reward scheme. The only information provided is the goal, which is embedded in the agent.

The second kind of apprentice agent uses learning by observation to acquire the knowledge of climbing the mountain. The agent shares the same constituents and capabilities of the experts it observes, being made out of a single part with two sensors and one actuator. The sensors provide the agent with its location in the mountain and its current speed (which can be positive if the agent is moving forwards or negative if the agent is moving backwards). The actuator provides the agent with three actions, accelerate forwards, accelerate backwards and maintain speed.

The following sections present the results of the simulations of these two scenarios. To provide the learning by observation agent with a broader set of experiences, all the simulations use more than one expert and each expert experiences the scenario in different ways, that is, they receive different information from the environment through their sensors. On the other hand, only one apprentice agent is used on the simulations. This simplification ensures that there is no risk for the tested apprentice agent to observe other apprentices, which may mislead it with incorrect actions, instead of observing the experts.

The unit of time used for the simulation results is the simulation step. A simulation step represents the time slot where the participant agents take a decision. For learning by observation agents, a simulation step can either represent an observation (the acquisition of a snapshot from the expert) and the subsequent learning (when the agent is in the learning state), or updating the facing conditions (see section 3.1), proposing and executing the action best suited for the faced conditions (when the agent is in the execution state) (see section 4). For reinforcement learning agents a simulation step represents updating the facing conditions, selecting an action for those conditions (from the action-state pairs stored in their memories) executing the action and interpreting the rewards (updating the action-state pairs). For expert agents, a simulation step represents updating the facing conditions and selecting and executing the action best suited for those conditions.

## 5.1 Results of the Virtual Hand Scenario

This section shows the impact of changing the confidence thresholds (see section 4.3) on the time spent on learning and on the total number actions that are appropriately executed by the agent. It also shows the way the agents perform, in terms of the number of correct actions, in two different settings.

The two settings depend on the sequence of numbers provided by the number generators. To control the results of the simulation, the numbers provided by the generators are designated from a pre-determined sequence. The number generators provide the numbers following that sequence and when the end of the sequence is reached the source needs to be reset by the agent. The reset makes the number generator provide the numbers following the same sequence.

For the first setting (*exp1*) the number generators of both apprentice and expert agents provide the same sequence of numbers. In this setting the apprentice agent faces the same conditions in the same sequence as the experts it observes which provides a good testing ground for the recall method of learning (see section 4.2). The setting shows how the agent is capable of performing the task as it was observed on the experts.

On the second setting (*exp2*), each number generator has a different sequence of numbers with different sizes, that is, the number generators need to be reset at different times. The apprentice agent faces conditions that are different from those faced by the observed experts which provides a good testing ground for the classification method of learning (see section 4.2). The setting shows how the agent is capable of performing a task that is similar to what has been observed when facing different conditions.

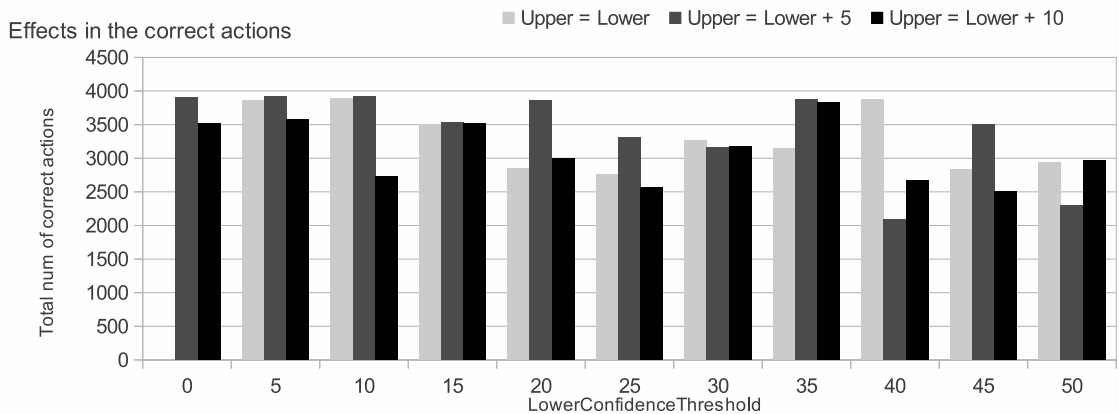

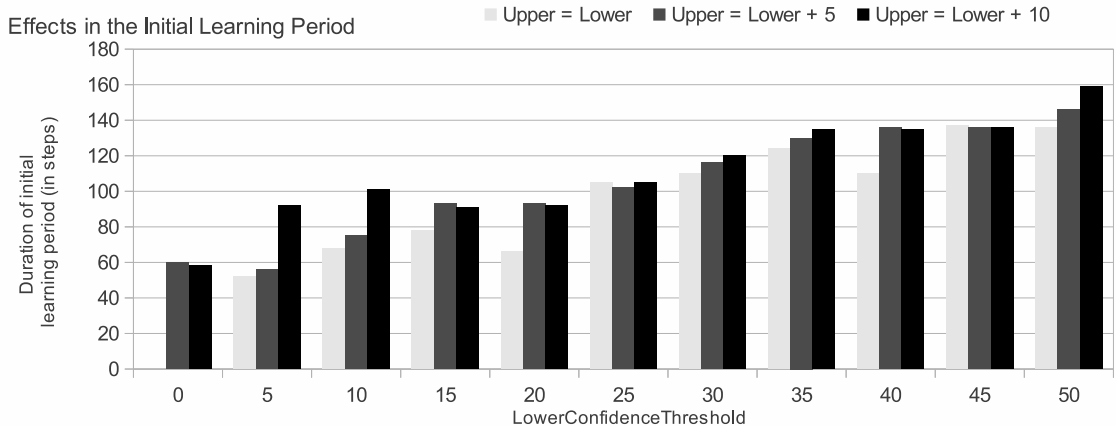

Figure 8: The impact of changing the confidence thresholds

Besides being a testing ground for the two learning methods, this scenario also determines the impact of changing the values for the UpperConfidenceThreshold and the LowerConfidenceThreshold (see section 4.3). For this reason, both settings were initially tested on agents using different values for these thresholds. Figure 8 presents a summary of the results obtained for the second setting in simulations with 4000 steps (each variation of the thresholds is tested in a simulation lasting 4000 steps). The choice for the second setting in Figure 8 is because it is where changing the thresholds had more influence and also because it is a more realistic approach for the scenario, since the chances for the agent to find itself on the same situation as the experts it observes are very small.

As explained in section 4.3, the confidence thresholds affect the length of the learning period and the number of subsequent learning periods which in turn affects the total number of correct actions performed by the agent throughout the simulation. When the UpperConfidenceThreshold is low, the agent may not have enough time to learn, which decreases the agent's performance in terms of being able to execute the correct actions. When the UpperConfidenceThreshold is too high the agent spends a lot of time learning, which reduces the time spent in the execution state and therefore the total number of actions executed by the agent is lower.

When the LowerConfidenceThreshold is too close to the UpperConfidenceThreshold the agent switches between the learning and execution state more often (see section 4). This may hinder the agent's ability to complete a task because the slightest error causes the agent to return to the learning state. When the LowerConfidenceThreshold is too far apart from the UpperConfidenceThreshold, the agent takes longer to switch between learning and execution. This slows down the ability to recognize the mistakes and switching to the learning state. It also slows down the recovery from the learning state to the execution state.

The results presented in Figure 8 fall in the region where the LowerConfidenceThreshold changes between zero and fifty and the UpperConfidenceThreshold changes from being equal to the LowerConfidenceThreshold and up to a difference of ten units more than the LowerConfidenceThreshold. This is where the agent is able to perform the maximum number of correct actions in the time provided by the simulation (4000 steps). If the difference between the thresholds is larger than ten, the total number of correct actions decreases since the agent takes longer to change between the learning and execution states.

Figure 8 also shows that the length of the initial learning period increases as the LowerConfidenceThreshold increases. The longer the agent spends learning the less time it has to perform the actions. Therefore, as Figure 8 shows, the optimal values (for this scenario) for the thresholds are ten for the LowerConfidenceThreshold and fifteen for the UpperConfidenceThreshold. This provides the agent with a learning period that is long enough for the agent to learn all the necessary skills (so that the majority of the actions it performs are correct) but short enough to allow the agent to perform the maximum amount of actions (which is 3923 actions according to Figure 8).

After determining the best values for the confidence thresholds, the scenario was simulated under the two settings (*exp1* and *exp2*) in a 4000 step simulation which was repeated 100 times to encompass the time variations that might exist on the simulations. Since the two settings were prepared with the two learning methods in mind, it is important to look at the importance given by the agent to these methods on each setting. Table 1 shows the average values of the weights of the recall and classification methods of learning (see section 4.2) on each setting.

Table 1 shows that, as expected, the recall method has more influence on the proposed actions when the agent is faced with the same conditions as the expert (*exp1*) than when it faces different conditions (*exp2*). In the first setting, the weight of the recall method makes it more likely (more than 50 % chance) for the actions proposed by this method to be executed, even though the actions proposed by the classification method can also be executed. In the second setting, the classification method has more influence on the proposed actions because it has the largest weight. The agent is not able to use the recall method very often to follow the same sequence of actions as the expert because it is facing conditions that are different from those observed on the expert (see section 4.2).

| Setting | Recall Weight | Classification Weight |
|---------|---------------|------------------------|
| exp1 | 0.509 | 0.491 |
| exp2 | 0.317 | 0.683 |

Table 1: The average weights of the recall and classification methods on each setting

The overview of the results of simulating the scenario in the two settings is presented in Table 2. The table compares the time spent on the learning state and on the execution state, the number of actions the agent was able to execute, how many of those actions were appropriate and also the time of a simulation step when learning and when executing actions. The results in Table 2 present both the average values and the standard deviation from running the simulation 100 times.

| | | Expert | Apprentice (exp1) | Apprentice (exp2) |
|---|---|--------|-------------------|-------------------|
| Time spent | average | - | 1.62 | 134.98 |
| learning (s) | stdev | - | 0.12 | 22.503 |
| Time spent | average | 2.415 | 10.435 | 23.365 |
| execution (s) | stdev | 0.606 | 0.957 | 6.017 |
| Total actions | average | 4000 | 3905 | 3709.25 |
| executed | stdev | 0 | 0 | 31.13 |
| Amount of | average | 100 % | 100 % | 92.95 % |
| appropriate actions | stdev | 0 | 0 | 7.73 p.p. |
| Step time in | average | - | 17.052 | 35.834 |
| learning state | stdev | - | 1.305 | 2.129 |
| Step time in | average | 0.603 | 2.672 | 6.363 |
| execution state | stdev | 0.151 | 0.245 | 1.815 |

Table 2: Overview of the results on the simulation of the virtual hand scenario

The results in Table 2 show that when the agent faces the same conditions as the experts it observes (*exp1*) it spends less time in the learning state and is able to perform more actions throughout the simulation than when it faces different conditions (*exp2*). When the agent faces different conditions (*exp2*) it spends more time learning because it needs to acquire more knowledge and requires this knowledge to come from different sources to improve the generalization in the classification method (see section 4.2). This leads the agent to spend more time discovering different experts to learn from and also causes the agent to return to the learning state more often.

In addition, the high standard deviation values show that when the agent faces different conditions (*exp2*) the number of actions executed throughout the simulation, the time spent learning and the time spent executing actions differs considerably. This variation is directly associated with the randomness of the observed experts. Each time the simulation is run, the apprentice observes different subsets of the available experts in a different sequence, which changes the knowledge retained by the agent throughout the simulation and affects the number of actions executed and the time spent on learning and on executing actions.

Table 2 also shows that when facing the same conditions (*exp1*) all the actions executed by the agent were appropriate, whereas when facing different conditions (*exp2*) only approximately 92% of the totality of executed actions were appropriate. This is an important indicator that the agent is more likely to return to the learning state, after starting to execute actions, when it is facing different conditions (*exp2*).

As for the average time spent in the simulation steps, Table 2 shows that in both cases, the simulation step is longer when agents are learning. This was expected since when the agent is in the learning state it has to perform various tasks such as discovering expert agents, comparing its software image with the software images of the discovered experts, acquiring the snapshots and storing its information, proposing actions for the conditions from the snapshots and evaluating those proposals (see section 4).

When executing actions, the simulation step of the apprentice agents is longer than the simulation step of the experts, which is understandable given that apprentice agents require more processing than the expert. Besides using two methods for proposing actions it is also necessary to take into account the influence of the internal evaluation. The step time while executing actions is also longer when the apprentice agent is facing different conditions (*exp2*), which influences the time spent executing actions. Even though the apprentice agent executes fewer actions (only 3709), when compared with when facing the same conditions (3905), it spends more time executing those fewer actions. The same effect is observed in the time spent by the simulation step when learning.

This happens because when facing different conditions the agent needs to acquire more knowledge which increases the amount of information in the agent's memory. Since the recall and classification methods of learning need to process the information contained in memory, the larger the amount of information the longer it takes to process it. This effect is felt both when learning and when executing actions because the agent uses these methods in both cases.

A closer look at the simulation results is presented in Figure 9, which shows the progress of the apprentice agents, in terms of the number of correct actions they have executed, throughout the simulation. The figure also shows the progress of the state of the learning process (see section 4) throughout the simulation in the apprentice agent. To provide a clearer presentation, the results are combined in groups of 100 simulation steps.

Figure 9 shows that, when apprentice agents face the same conditions as the expert (*exp1*) after a short learning period (of about 200 steps) all the actions they execute are correct. When the agent faces different conditions from the experts it observes (*exp2*), the initial learning period lasts a little longer (about 300 steps) and only approximately 90% of the executed actions are correct.

The agent also experiences additional learning periods (of short duration) throughout the simulation. The additional learning periods are mainly caused by the evaluation activity mechanism that forces apprentices to switch to the learning state when facing conditions that were not observed (see section 4.3). Although the subsequent learning periods affect the time spent on learning they are of short duration. As Figure 9 shows, the overall permanence in these subsequent learning periods does not exceed more than 3% of the total simulation steps.

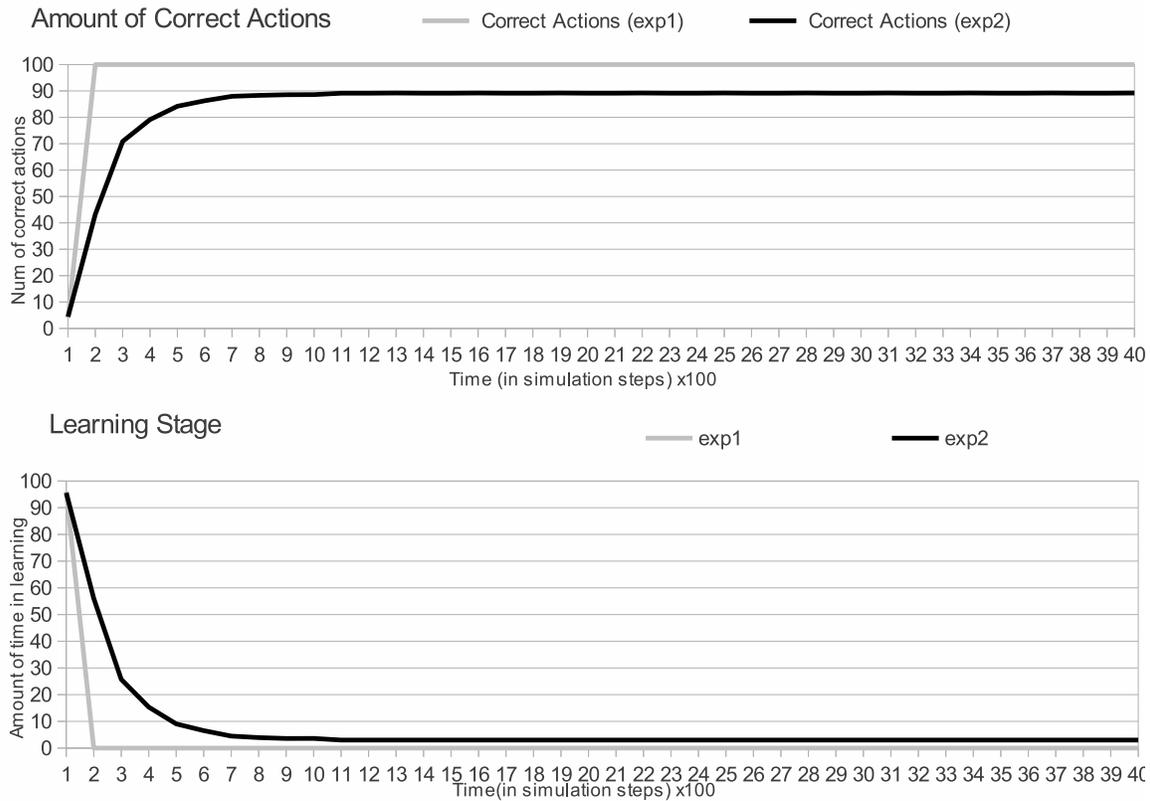

Figure 9: The behaviour of the apprentice agent in the two settings throughout the simulation

## 5.2 Results of the Mountain Car Scenario

The mountain car scenario compares expert agents, which are specialized in climbing mountains, with the learning by observation agents (LbO) that learn by observing experts performing the task to learn and the reinforcement learning agents (RL) that learn through reinforcements. All the participant agents face the same conditions, that is, they are placed in the same mountain at the same starting points. The observed experts also face the same conditions as the learning by observation agent.

The results present a comparison based on the number of actions, the distance travelled and the time spent until reaching the top of the mountain (the goal of the simulation). The results also determine the average time of a simulation step for each agent and how much time it takes for an agent to learn, that is, how much time it takes for an agent to reach the top of the mountain for the first time.

Since this scenario provides a goal, the simulation time-frame is grouped by attempts of achieving the goal. Each attempt lasts a variable number of simulation steps, with a maximum duration of 500 simulation steps. If the agent achieves the goal before the 500 steps the attempt is completed and is regarded as successful. If after the 500 steps the goal is not achieved, the attempt is regarded as failed. The simulation lasts for 50000 attempts, to provide the reinforcement learning agents enough time for testing all the hypotheses and provide their best results.

Table 3 presents a summary of the important aspects of the scenario such as the amount of time, number of attempts, number of actions and distance travelled by each agent to reach the top for the first time. The table also shows how many times the agent reached the top of the mountain, the average time of a simulation step and the average amount of simulation steps in an attempt to reach the goal. The data in Table 3 was obtained from running the scenario 100 times to encompass time variations. The student's t-test was used to ensure the statistical relevance of the data collected from running the scenario.

| | Expert | Apprentice (LbO) | Apprentice (RL) | T-TEST (LbO - RL) |
|---|---|---|---|---|
| Time to reach top first time (ns) | 30 | 3300 | 14660 | $2.2 \times 10^{-7}$ |
| Attempts for reaching top first time | 1 | 1 | 105 | $3.4 \times 10^{-10}$ |
| Actions executed to reach top first time | 136 | 278 | 52471 | $3.2 \times 10^{-10}$ |
| Distance travelled to reach top first time | 2.73 | 2.73 | 610.37 | $2.5 \times 10^{-9}$ |
| Number of times reached top | 50000 | 50000 | 49895 | $2.9 \times 10^{-10}$ |
| Average simulation step time (ms) | 0.28 | 4.94 | 0.31 | $2.7 \times 10^{-39}$ |
| Average simulation steps spent in an attempt | 136 | 136.01 | 229.92 | $4.9 \times 10^{-12}$ |

Table 3: Overview of the results on the simulation of the mountain car scenario

Table 3 shows that, as expected, the expert agent exhibits the best results in all aspects. The table also shows that the learning by observation agent outperforms the reinforcement learning agent in all aspects with the exception of the time of a simulation step. The learning by observation agent also gets close to the same results as the expert in all aspects with the exception of the time it takes to reach the top and the simulation step time.

When compared with the reinforcement learning agent, the simulation step of a learning by observation agent lasts longer because of the amount of processing behind that decision.

Nevertheless, the learning by observation agent requires less simulation steps to complete an attempt, that is, to reach the goal. The main reason for the longer simulation step is because the learning by observation agent uses two methods for proposing the actions (the recall and classification methods as explained in section 4.2) which requires an additional effort of combining the results and choosing the best action from them. The agent also has to perform a set of other tasks, like for example the internal evaluation (see section 4.3). The performance of the software used by the agent is likely to be improved in future versions, which would lead to an improvement in the simulation step time.

Although the simulation step of a reinforcement agent takes less time, the agent requires more steps and attempts to reach the top of the mountain for the first time and eventually takes more time to learn how to reach the top of the mountain as the results in Table 3 show. This puts the reinforcement agent in last place when considering the time it takes to reach the top for the first time. The results of the t-test in Table 3 show that the data acquired in the simulations is statistically relevant.

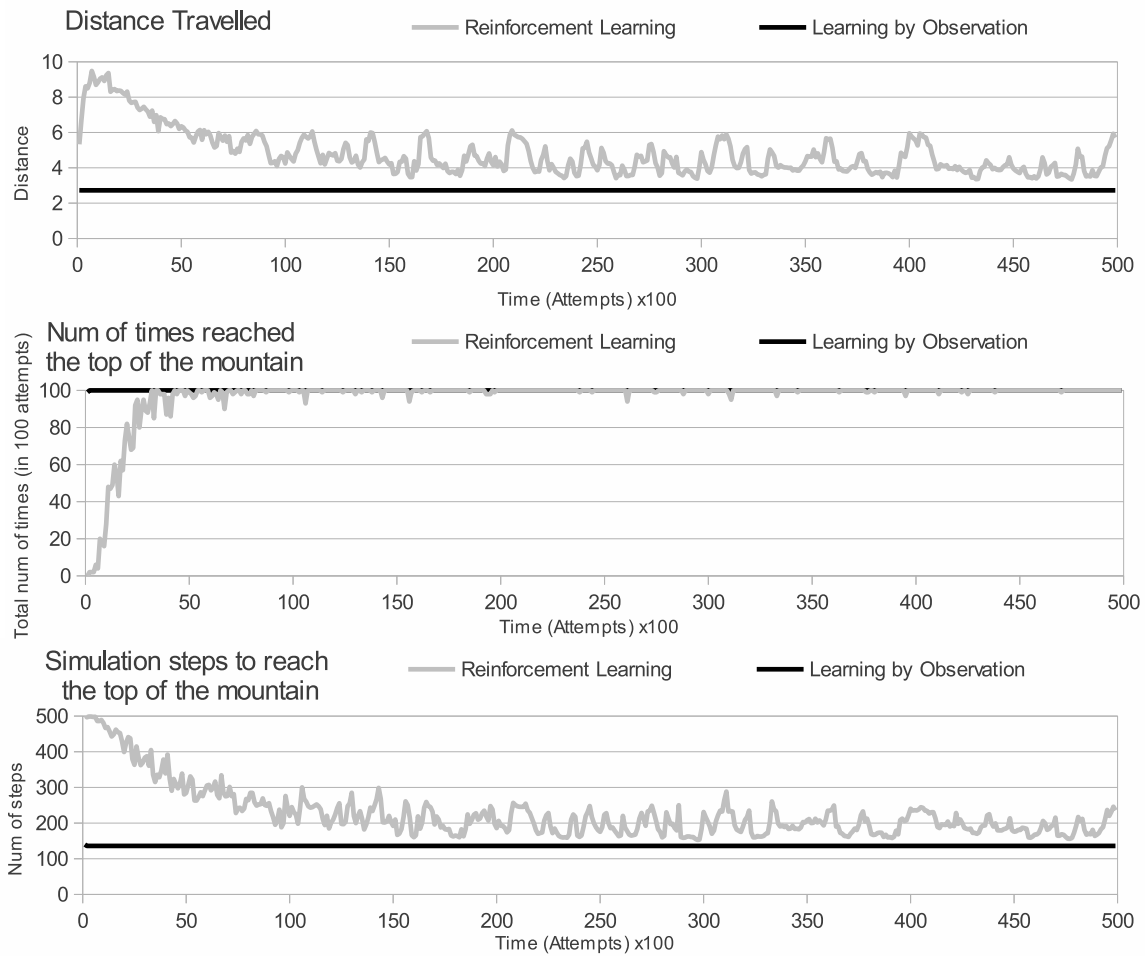

Figure 10: General view of the results for the second scenario

In addition to the information on achieving the goal for the first time, it is also important to know how the agents performed throughout the simulation. Figure 10 shows the progress of the expert and the two apprentice agents (learning by observation and reinforcement learning) throughout the simulation in terms of the distance travelled, the number of times reaching the top and the number of simulation steps required to reach the top. For a clearer presentation, the results are combined in groups of 100 attempts to reach the goal. Each attempt lasts a number of simulation steps that ranges from 136 to 500, depending on the number of steps that are necessary to reach the top of the mountain (see Table 3).

The results presented in Figure 10 show that the learning by observation agent performs better than the reinforcement learning agent in all considered dimensions. The figure shows that the learning by observation agent requires less simulation steps (in comparison with the reinforcement learning apprentice) to go from the starting point to the top of the mountain after learning how to do it. The simulation steps spent in each attempt stabilizes at 136 (which is the same number steps spent by the expert) right after the agent has learnt the task of climbing the mountain. The learning by observation agent also requires less attempts to learn how to reach the top (1 attempt as shown in Table 3).

The distance travelled by the learning by observation agent is also smaller than the distance travelled by the reinforcement learning agent. The value of the distance, in the learning by observation agent, stabilizes at 2.73 (the same as the expert as Table 3 shows) right after the agent learns the task of climbing the mountain, unlike what happens with the reinforcement learning agent. As Figure 10 shows, even after a long learning period, the reinforcement learning agent is not able to learn the most efficient way to climb the mountain (the one that requires the least amount of actions and the smallest distance). Even the lowest values of the fluctuations of the number of actions and the travelled distance are still far away from those obtained by the learning by observation agent.

Besides these inabilities, the reinforcement learning agent's capability of reaching the goal fluctuates between 90% and 100% of the times, until almost the end of the simulation. This means that even after learning how to reach the goal, the reinforcement learning agent is not always able to do it, something that is also observed in the total number of times the agent reaches the top of the mountain in Table 3. In contrast, the learning by observation agent exhibits more stable results after learning the task of reaching the top of the mountain.

## 6. Conclusions and Future Work

The adoption of learning by observation solutions is relatively new in computer science and, as the latest approaches show, it is still under development (Sullivan, 2011; Kulic et al., 2011; Tan, 2012; Fonooni, Hellström, & Janlert, 2012). With the exception of our work (Costa & Botelho, 2011, 2012) and Machado and Botelho's (2006) work, all the contributions for learning by observation are focused on robotic agents and their physical properties. Software agents and even the software used in robots are neglected. Even Machado and Botelho's work has limitations since it only addressed the problem of learning vocabulary and it does not allow generalizations of the acquired knowledge. This means that, unlike our approach, her apprentice agent is not able to learn control mechanisms neither is capable of dealing with conditions that are different from those observed on the expert.

Therefore, our learning approach clearly contributes to advance the state of the art on learning by observation, providing software agents with a learning solution that is different from all other methods that are usually applied for software agents. Unlike the robotic approaches for learning by observation, our software approach is not limited to software agents. It can also be used by robotic agents with minor adaptations, since all their physical actions are controlled by or reflected in software events. Even the data collected by robotic sensors needs a software representation (however complex it can be), since in its core the robot is effectively running a program and all high level decisions are made by that program. Thus, a software approach provides a broader solution than approaches limited to the physical properties of robotic environments.

The experimental results in section 5 show that software agents are capable of improving their ability to perform a task using our approach. The first scenario also shows that the classification method of learning is essential for situations where the agent is not facing exactly the same sequence of events as the expert. Namely, situations in which the agent is faced with conditions that are different from those faced by experts it has observed. The most usual scenarios for learning by observation, reported in the literature, address cases in which the apprentice and the expert agents face exactly the same conditions. This kind of situation is ideal for sequence learning, which is the reason why this method is one of the most used in learning by observation approaches. With the inclusion of the classification method, the ability to learn by observation is extended to other problems and domains.

The combination of the classification method with the recall method, which is inspired in sequence learning, ensures that the agent is able to adapt, by itself, to a larger number of circumstances, including those that are most usual for learning by observation scenarios. Besides the combination of these two methods, our approach also provides an internal evaluation mechanism that constantly tests the agent's knowledge. This allows the agent to know when it needs to acquire more knowledge or when it has acquired sufficient knowledge and therefore can execute actions. The approach also offers the possibility of using external feedbacks to enhance the agent's evaluation, which gives agents the ability of knowing if the actions they execute are appropriate.

The agent's internal evaluation enables learning by observation agents to enhance their knowledge even after they stop observing experts and start using the acquired knowledge, since after starting to use their knowledge they may go back to the learning state. This was one of the major drawbacks of previous learning approaches. The usual way of handling this is by manually feeding new examples whenever the agent requires them. In the case of our approach the agent is able to decide by itself when it needs to return to a learning state, to observe experts. The process does not require any intervention since the agent can find the experts from which to collect the new training examples.

The application of our approach in two distinct scenarios shows that it can be adapted to different domains. The scenarios also show that the amount of time taken by our learning approach to decide the actions to execute can sometimes be high. However, as the second scenario shows, our agents are able to achieve the goal in less time than a reinforcement learning approach, even in situations that have already been shown by the literature to be adequate for reinforcement learning. The learning by observation agents are also able to achieve approximately the same results as the experts they observe.

Although apprentice agents take more time to choose the actions to execute than the experts, this difference is smaller on the first scenario. Considering that, in the first scenario, the expert's knowledge is expressed with more rules than in the second scenario, our approach, unlike reinforcement learning, is able to cope with the increase of the complexity in terms of the number of rules required to express the expert's knowledge.

Besides providing a new insight for learning in software agents, through the software image, our approach can also contribute to the software embodiment problem. Although its main purpose is directed to learning by observation, the software image also allows visible software agents to represent themselves on what can be called a body. As future work we may continue developing the software image to better adapt it to the software embodiment problem.

We also intend to continue testing our approach in different scenarios, especially in situations where the actions have effects both in the agent and in the environment. In such situations only a part of the effect of the actions is visible in the environment. We expect the results to be similar to a situation where all the effects of the actions are visible in the environment when the visible effects are enough to distinguish the actions. However, when the visible effects are not enough to distinguish the actions, agents who learn only from the effects of the actions might not be able to learn properly.

For example, the effect of action one is removing a number from the environment and adding it to the agent's internal memory and the effect action two is removing a number from the environment and subtracting it to the agent's internal memory. Both actions have different effects in the agent, but the visible effects on the environment are exactly the same. In this situation, an agent that only learns from the visible effects of the actions will not be able to distinguish these two actions.

Finally, we also consider improving the approach to open the possibility of an agent performing a task that is radically different from the tasks performed by the observed experts. The agent should be able to use the acquired knowledge in a way that allows it to perform new tasks that were never observed before.

## Acknowledgments

This paper reports PhD research work, for the Doctoral Program on Information Science and Technology of ISCTE-Instituto Universitario de Lisboa. It is partially supported by Fundação para a Ciencia e a Tecnologia through the PhD Grant number SFRH/BD/44779/2008 and by FCT project PEst-OE/EEI/LA0008/2013